\documentclass[runningheads]{llncs}

\usepackage{eccv}

\usepackage{eccvabbrv}

\usepackage{graphicx}
\usepackage{booktabs}
\usepackage{caption}

\ifdefined\pdfcompresslevel\else\newcount\pdfcompresslevel\fi
\ifdefined\pdfoptionpdfminorversion\else\newcount\pdfoptionpdfminorversion\fi
\ifdefined\pdfgentounicode\else\newcount\pdfgentounicode\fi
\ifdefined\pdfglyphtounicode\else\newcommand{\pdfglyphtounicode}[2]{}\fi
\usepackage[accsupp]{axessibility}  

\usepackage{graphicx}
\usepackage{tikz-cd}
\usepackage{tikz}
\usetikzlibrary{positioning, arrows.meta, fit, calc, backgrounds}
\usepackage{multicol}
\usepackage{enumitem}
\usepackage{siunitx}

\usepackage{hyperref}

\usepackage{orcidlink}

\usepackage{url}

\usepackage{enumitem}

\makeatletter
\def\@maketitle{\newpage
 \markboth{}{}%
 \def\lastand{\ifnum\value{@inst}=2\relax
                 \unskip{} \andname\
              \else
                 \unskip \lastandname\
              \fi}%
 \def\and{\stepcounter{@auth}\relax
          \ifnum\value{@auth}=\value{@inst}%
             \lastand
          \else
             \unskip,
          \fi}%
 \begin{center}%
 \let\newline\\
 \vspace*{-8mm}%
 {\Large \bfseries\boldmath
  \pretolerance=10000
  \@title \par}\vskip .45cm
\if!\@subtitle!\else {\large \bfseries\boldmath
  \vskip -.65cm
  \pretolerance=10000
  \@subtitle \par}\vskip .8cm\fi
 \setbox0=\vbox{\setcounter{@auth}{1}\def\and{\stepcounter{@auth}}%
 \def\thanks##1{}\@author}%
 \global\value{@inst}=\value{@auth}%
 \global\value{auco}=\value{@auth}%
 \setcounter{@auth}{1}%
{\lineskip .5em
\noindent\ignorespaces
\@author\vskip.3cm}
 {\small\institutename}
 \end{center}%
 }
\makeatother

\begin{document}

\title{Geometrically Consistent Multi-View Scene Generation from Freehand Sketches} 

\titlerunning{S2MV}

\vspace{-7mm}
\author{Ahmed Bourouis\inst{1,2}\orcidlink{0009-0007-8827-387X} \and
Savas Ozkan\inst{1}\orcidlink{0009-0006-0450-0645} \and
Andrea Maracani\inst{1}\orcidlink{0000-0002-6217-8731} \and
Yi-Zhe Song\inst{2}\orcidlink{0000-0001-5908-3275} \and
Mete Ozay\inst{1}\orcidlink{0000-0002-2515-1062} }

\authorrunning{Bourouis et al.}

\vspace{-2mm}

\institute{$^1$Samsung Research, UK \quad $^2$SketchX, University of Surrey, UK}

\maketitle

{\centering
\includegraphics[width=\textwidth]{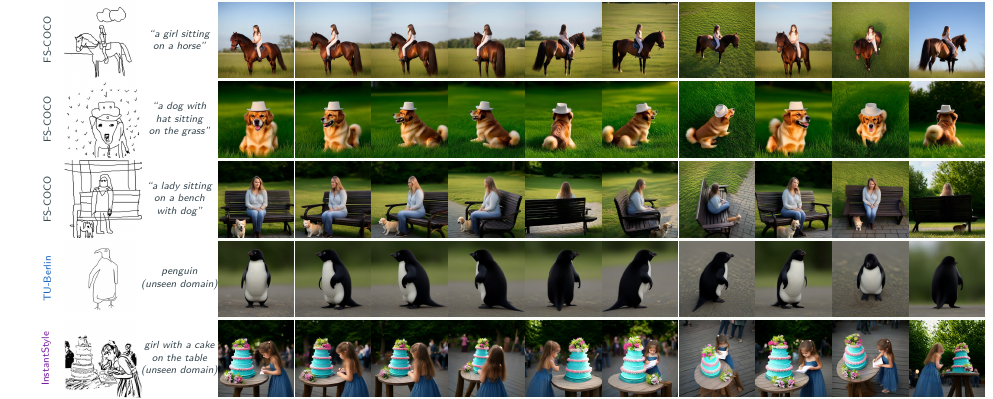}
\captionof{figure}{\textbf{Sketch-to-Multi-View generation.} Given a single freehand sketch and a text caption, our method generates geometrically consistent multi-view images spanning a full $360^{\circ}$ azimuth orbit at four different elevations. Top three rows: in-domain examples from FS-COCO~\cite{chowdhury2022fs}. Bottom two rows: zero-shot generalisation to unseen sketch domains (TU-Berlin~\cite{eitz2012humans} and InstantStyle~\cite{tang2025instance}). \href{https://ahmedbourouis.github.io/S2MV/}{\texttt{Project Page}}.}
\label{fig:teaser}
\par}
\vspace{-4mm}

\begin{abstract}
We tackle a new problem: generating geometrically consistent multi-view scenes from a single freehand sketch. Freehand sketches are the most geometrically impoverished input one could offer a multi-view generator. They convey scene intent through abstract strokes while introducing spatial distortions that actively conflict with any consistent 3D interpretation. No prior method attempts this; existing multi-view approaches require photographs or text, while sketch-to-3D methods need multiple views or costly per-scene optimisation.

We address three compounding challenges; absent training data, the need for geometric reasoning from distorted 2D input, and cross-view consistency, through three mutually reinforcing contributions:
(i) a curated dataset of $\sim$9k sketch-to-multiview samples, constructed via an automated generation and filtering pipeline; (ii) Parallel Camera-Aware Attention Adapters (CA3) that inject geometric inductive biases into the video transformer; and (iii) a Sparse Correspondence Supervision Loss (CSL) derived from Structure-from-Motion reconstructions.

Our framework synthesizes all views in a single denoising process without requiring reference images, iterative refinement, or per-scene optimization. Our approach significantly outperforms state-of-the-art two-stage baselines, improving realism (FID) by over 60\% and geometric consistency (Corr-Acc) by 23\%, while providing up to a 3.7$\times$ inference speedup.

\vspace{-2mm}
\vspace{-3mm}

\end{abstract}

\section{Introduction}
\label{sec:intro}

Consider a rough, freehand sketch of a street scene: a few strokes for a car, some wobbly lines representing buildings, a stick figure on the pavement. Despite this inherent abstraction, a human observer effortlessly infers a plausible 3D world, understanding that the car has an occluded side, the buildings extend beyond the visible corners and the figure could be seen from different perspectives. This rises the fundamental question: 

\begin{quote}
\textit{Can a generative model, conditioned solely on a single freehand sketch, synthesize a set of geometrically consistent novel views representative of a complete 3D scene?}
\end{quote}

We define this task as \textbf{Sketch-to-Multi-View} (\textbf{S2MV}) generation as illustrated in Fig.~\ref{fig:teaser}. This task is a surprisingly hard problem for current generative models, and one that has not been studied before. Freehand scene sketches are arguably the most geometrically impoverished conditioning signal one could provide to multiview generators. Unlike photographs, which provide dense appearance and implicit 3D cues (i.e., shading and perspective), or text prompts, which offer semantic flexibility without spatial constrains, freehand sketches are \emph{simultaneously} spatially informative yet geometrically unreliable. They convey the \emph{intent} of a scene layout while introducing abstraction, non-linear distortions, and stroke-level noise that often conflict with a consistent 3D interpretation.

Existing multi-view diffusion methods do not directly address these inconsistencies. Image-conditioned approaches~\cite{liu2023zero,shi2023zero123++,liu2023syncdreamer,shi2023mvdream,voleti2024sv3d,gao2024cat3d,zhao2025hunyuan3d} typically begin with high-fidelity photographs that already resolve geometric ambiguity. Text-conditioned methods~\cite{shi2023mvdream} operate in a domain where spatial imprecision is expected and tolerated. Conversely, existing Sketch-to-3D methods, such as Sketch2NeRF~\cite{chen2024sketch2nerf} and SketchDream~\cite{liu2024sketchdream}, are largely restricted to single-object reconstruction or require \emph{multiple} sketch views and costly per-scene optimisation. To our knowledge, no prior work generates geometrically consistent multi-view outputs from a single, freehand scene sketch.


Bridging this gap requires solving three compounding challenges. \textbf{(i)~Data:} no sketch-to-multi-view dataset exists, and freehand sketches cannot be consistently paired with multi-view images through standard rendering pipelines. \textbf{(ii)~Geometric reasoning:} generative models must infer coherent 3D structures from a 2D input that lacks strict projective alignment, demanding camera-aware generation mechanisms. \textbf{(iii)~Cross-view consistency:} generating different views simultaneously requires the model to enforce correspondences \emph{between} views, rather than simply conditioning on viewpoint independently.

We address these challenges with the following contributions:
\begin{itemize}
\item \textbf{A Curation Pipeline for S2MV Data} (\cref{sec:method:data}). We introduce a novel pipeline to bridge the domain gap between freehand sketches and multi-view images. By leveraging FS-COCO~\cite{chowdhury2022fs}, we synthesize photorealistic frontal views using a pretrained generative model~\cite{flux-2-2025}, verify sketch-image alignment through semantic segmentation matching~\cite{bourouis2024open,carion2025sam}, and generate multiple views per scene~\cite{dx8152_qwen_edit_2509_multiple_angles_2025}. The resulting dataset includes 9\,222 curated samples with $33$ views each, providing a rigorous foundation for S2MV training.

\item \textbf{Camera-Aware Attention Adapters (CA3)} (\cref{sec:method:camera}). We propose a lightweight camera attention adapters that inject Projective Rotary Position Encoding (PRoPE)~\cite{li2025cameras} into a pretrained video diffusion transformer~\cite{wan2025wan}. This mechanism adds only 2.7\% additional parameters while enabling attention layers to explicitly account for relative projective geometry between views.

\item \textbf{Correspondance Supervision Loss (CSL)} (\cref{sec:method:corr}). We introduce a sparse InfoNCE-based correspondence loss applied to the adapter's latent projections to maintain geometric consistency between cross-views. Using positive pairs extracted through Structure-from-Motion~\cite{schonberger2016structure},this regularization enforces cross-view identity and merges distinct viewpoints into a unified latent representation.



\end{itemize}

The components of our framework are mutually reinforcing, creating a cohesive system for S2MV synthesis. While our curated dataset provides the foundational sketch--multiview pairs previously missing from the literature, our CA3 provides the necessary geometric inductive bias and the \emph{capacity} to reason about viewpoints. Last but not least, our CSL provides the explicit supervisory signal that teaches the model \emph{which} spatial locations across views should attend to each other, turning geometric capacity into geometric consistency. Removing any one of these components significantly degrades the final performance.

Crucially, our method generates all multi-views jointly in a \emph{single end-to-end denoising process} directly from a sketch, bypassing any intermediate photo-generation stage, requiring no reference photograph, no per-scene optimisation, no iterative refinement. To the best of our knowledge, this is the first method to produce geometrically coherent, photorealistic multi-view scenes from freehand sketch input.

\section{Related Work}
\label{sec:related}

\noindent\textbf{Multi-View Generation and Camera Conditioning.}
Diffusion models have been widely applied to multi-view image generation.
Early methods fine-tune image diffusion models to synthesise novel views from a single photograph: Zero-1-to-3~\cite{liu2023zero} conditions on relative camera pose, and follow-up works~\cite{shi2023zero123++,liu2023syncdreamer,shi2023mvdream,li2024era3d,long2024wonder3d} improve consistency by generating multiple views jointly through synchronised diffusion or multi-view attention.
A parallel line of work repurposes video diffusion models, treating the temporal axis as a viewpoint axis, for object-centric orbiting~\cite{voleti2024sv3d,chen2024v3d} or scene-level~\cite{gao2024cat3d} generation.
More recent DiT-based architectures push the scale further: Hunyuan3D\,2.0~\cite{zhao2025hunyuan3d} trains a 2B-parameter shape-generation DiT, TRELLIS~\cite{xiang2025native} employs rectified flow transformers over structured 3D latents, and See3D~\cite{ma2025you} learns 3D priors from 16M video clips without pose annotations.

How camera geometry is communicated to the generative model is a critical design axis.
Absolute conditioning methods~\cite{liu2023zero,he2024cameractrl,wang2024motionctrl} inject camera parameters as global vectors, while per-pixel Pl\"ucker ray representations~\cite{he2024cameractrl,xu2024camco,zhou2025stable} offer richer spatial encoding but do not inject \emph{relative} inter-view geometry into the attention computation.
Geometric approaches address this limitation directly: PRoPE~\cite{li2025cameras} modulates queries, keys, and values with projective matrices so that attention scores reflect relative camera geometry, UCPE~\cite{zhang2025unified} extends this to per-token ray-level encoding, and EpiDiff~\cite{huang2024epidiff} and SPAD~\cite{kant2024spad} restrict cross-view attention to epipolar lines.
AC3D~\cite{bahmani2025ac3d} shows that camera information concentrates in a subset of DiT layers, enabling a $4{\times}$ parameter reduction, and the concurrent ReRoPE~\cite{li2026rerope} injects relative pose into underutilised RoPE frequency bands.

We build on the Wan\,2.1 video DiT~\cite{wan2025wan} and adopt PRoPE inside lightweight parallel adapters (adding fewer than 3\% parameters) following the zero-initialised parallel branch paradigm~\cite{zhang2023adding}. We complement the geometric inductive bias with explicit correspondence supervision, a combination not explored in prior work.

\noindent\textbf{Geometric Consistency and Correspondence Supervision.}
Ensuring cross-view geometric consistency in generated images remains a central challenge.
Implicit approaches encode geometric priors through architecture: SyncDreamer~\cite{liu2023syncdreamer} constructs a shared 3D feature volume, epipolar attention~\cite{huang2024epidiff,li2024era3d,tang2023emergent} constrains cross-view attention to geometrically plausible regions, and SpatialCrafter~\cite{zhang2025spatialcrafter} decouples spatial and temporal attention for scene reconstruction.
ViewCrafter~\cite{yu2024viewcrafter} guides video diffusion with monocular depth-based point clouds, and GeoVideo~\cite{bai2025geovideo} aligns predicted depth maps across frames during training.
In all these cases, geometric reasoning is an emergent property of the architecture rather than an explicitly supervised objective.

In the reconstruction domain, DUSt3R~\cite{wang2024dust3r}, MASt3R~\cite{leroy2024grounding}, and VGGT~\cite{wang2025vggt} learn to predict dense correspondences from multi-view images, but treat them as the end product rather than as supervision for generation.
CorrespondentDream~\cite{kim2024enhancing} extracts correspondences from a frozen diffusion UNet to regularise NeRF optimisation, yet does not train the generative model itself with correspondence supervision.

The concurrent CAMEO~\cite{kwon2025cameo} is closest to our work, supervising attention maps with VGGT-derived correspondences via dense cross-entropy on the main self-attention of a UNet-based model.
Beyond the apparent ``sparse-vs-dense'' contrast, our approach differs along four \emph{causal} axes.
(i)~Adapter vs.\ main attention: CAMEO supervises a self-attention layer of CAT3D's~\cite{gao2024cat3d} UNet, entangling correspondence with appearance on the \emph{same} parameters that solve denoising; we supervise a \emph{parallel}, zero-initialised adapter whose PRoPE-modulated queries/keys live in the relative-camera projective space, so geometric supervision never competes with appearance synthesis.
(ii)~Sparse contrastive vs.\ dense cross-entropy at video scale: CAMEO's dense map is tractable on $F{=}4$ views at $32{\times}32$ ($\approx$16M entries/layer); at our video-DiT scale ($4N{+}1{=}133$ frames, $60{\times}60$ latent) it would hold $\approx\!2.3{\times}10^{11}$ entries (hundreds of GB/layer), whereas our sparse InfoNCE~\cite{oord2018representation} costs ${\sim}$1\,MB/layer and explicitly repels negatives.
(iii)~Classical vs.\ learned correspondence: our SfM~\cite{schonberger2016structure} supervision carries no learned prior, whereas CAMEO inherits VGGT's pointmap biases.
(iv)~Video DiT vs.\ image backbone: CAMEO is validated on image backbones at $F{=}4$, while we show the supervision transfers to a video DiT generating $4N{+}1$ frames in a \emph{single} pass, the regime where dense supervision is infeasible.

\noindent\textbf{Sketch-Conditioned Generation.}
Sketch-based image generation has been widely studied through conditional diffusion models.
ControlNet~\cite{zhang2023adding} and T2I-Adapter~\cite{mou2024t2i} accept edge maps as spatial conditions but target clean line drawings rather than freehand sketches.
Recent work addresses the sketch abstraction gap: Koley~\etal~\cite{koley2024s} handle amateur sketches for single-image generation, and KnobGen~\cite{navard2024knobgen} adapts to varying sketch complexity.
SketchingReality~\cite{bourouis2026sketchingreality} generates photorealistic images from freehand \emph{scene} sketches using a CLIP-based sketch encoder, but is limited to single-view output.
In the 3D domain, Sketch2NeRF~\cite{chen2024sketch2nerf} requires multiple sketch views and optimises a NeRF via SDS, while SketchDream~\cite{liu2024sketchdream} combines sketch conditioning with depth guidance, both target single objects and require costly per-scene optimisation.

Our method extends the scene-sketch paradigm of~\cite{bourouis2026sketchingreality} to \emph{multi-view} output: from a single freehand scene sketch, we generate $N$ geometrically consistent views through a video diffusion transformer in a single forward pass, with no per-scene optimisation.

\section{Method}
\label{sec:method}

\definecolor{ourbg}{HTML}{E8F5E9}
\definecolor{ourbdr}{HTML}{388E3C}
\definecolor{fzbg}{HTML}{F5F5F5}
\definecolor{fzbdr}{HTML}{BDBDBD}
\definecolor{lossbg}{HTML}{FFEBEE}
\definecolor{lossbdr}{HTML}{C62828}
\definecolor{datbg}{HTML}{E3F2FD}
\definecolor{datbdr}{HTML}{1976D2}
\definecolor{txtclr}{HTML}{7B1FA2}
\definecolor{arrc}{HTML}{546E7A}
\definecolor{trainbg}{HTML}{FFF8E1}
\definecolor{trainbdr}{HTML}{F9A825}
\begin{figure*}[t]
  \centering
  \includegraphics[width=\textwidth]{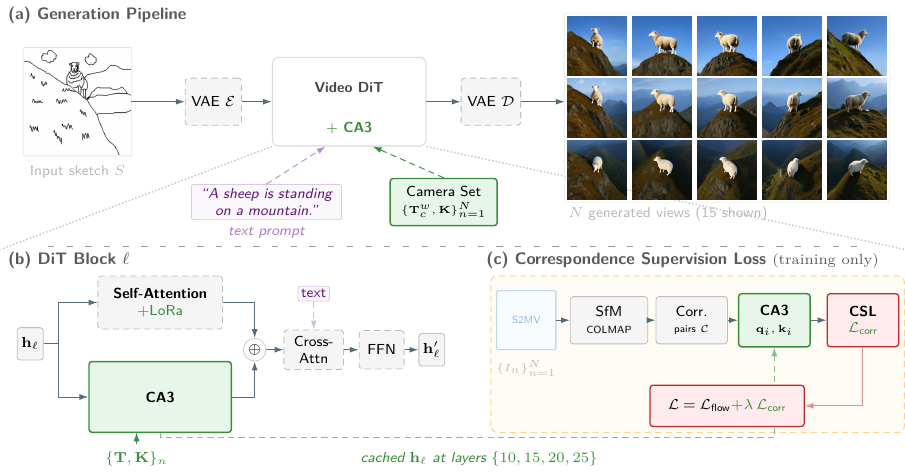}
  \caption{\textbf{Method overview.}
  (a)~A freehand sketch is encoded then denoised by a video DiT augmented with our \textcolor{ourbdr}{\textbf{CA3}}. All $N$ views are generated in a single forward denoising process.
  (b)~Each DiT block contains a self-attention with LoRa and a parallel CA3 to inject relative camera geometry.
  (c)~During training, 
  SfM correspondences from pseudo ground-truth views supervise the CA3 query--key projections via a sparse InfoNCE loss.
  Dashed boxes: frozen modules.
  \vspace{-4mm}
  }
  \label{fig:method_overview}
\end{figure*}

Given a freehand sketch $S$ and a set of $N$ target camera poses, our method generates $N$ geometrically consistent photorealistic views through a video diffusion transformer (\cref{fig:method_overview}.a) augmented with lightweight camera-aware adapters \textbf{CA3} (\cref{fig:method_overview}.b) and trained with correspondence supervision \textbf{CSL}(\cref{fig:method_overview}.c).

\textbf{S2MV} requires solving three compounding challenges: the absence of paired training data, the need for camera-aware generation, and the enforcement of cross-view geometric consistency. In this section, we first review the video diffusion backbone (\cref{sec:method:prelim}), followed by our dataset construction pipeline (\cref{sec:method:data}), Camera-Aware Attention Adapters (CA3) (\cref{sec:method:camera}), and Correspondence Supervision Loss (CSL) (\cref{sec:method:corr}).

\subsection{Preliminaries: Video Diffusion Transformers}
\label{sec:method:prelim}

Our framework is built upon a standard Video Diffusion Transformer (Wan~2.1~\cite{wan2025wan}), which is repurposed to support the multi-view sequence modeling required for S2MV. The model operates in a compressed latent space $\mathcal{Z}$ learned by a VAE encoder $\mathcal{E}$, which first maps a sequence of frames $[\mathbf{x}_i]_{i=0}^N$ into a latent representation $[\mathbf{z}_i]_{i=0}^N$. Later, the DiT architecture processes each $\mathbf{z}_i$ by patchifying it into a sequence of $T$ tokens whose dimension is $D$. Each of the $L$ transformer blocks computes self-attention as:
\begin{equation}
    \text{Attn}(\mathbf{Q}, \mathbf{K}, \mathbf{V}) = \text{softmax}\left(\frac{\mathbf{Q}\mathbf{K}^\top}{\sqrt{d}}\right)\mathbf{V},
\end{equation}
\noindent where $\mathbf{Q}, \mathbf{K}, \mathbf{V} $ are the query, key, and value projections. The training is performed via flow matching~\cite{lipman2022flow} by minimizing the velocity field error:
\begin{equation}
    \mathcal{L}_{\text{flow}} = \mathbb{E}_{t, \mathbf{z}_0, \boldsymbol{\epsilon}}\left[\|\mathbf{v}_\theta(\mathbf{z}_t, t, \mathbf{c}) - (\mathbf{z}_0 - \boldsymbol{\epsilon})\|^2\right],
\end{equation}
\noindent where $\mathbf{z}_t$ is the noisy latent at timestep $t \in [0,1]$, $\boldsymbol{\epsilon} \sim \mathcal{N}(\mathbf{0}, \mathbf{I})$, and $\mathbf{c}$ denotes the conditioning signals. 
The VAE applies $4{\times}$ temporal compression, effective for video where consecutive frames are redundant, but destructive for multi-view generation where neighbouring views differ substantially. We therefore replicate each view four times before encoding (\emph{frame replication}), yielding $4N{+}1$ input frames and exactly one latent per view.

\subsection{Sketch-to-Multiview Dataset Construction}
\label{sec:method:data}

\begin{figure*}[t]
  \centering
  \includegraphics[width=\textwidth]{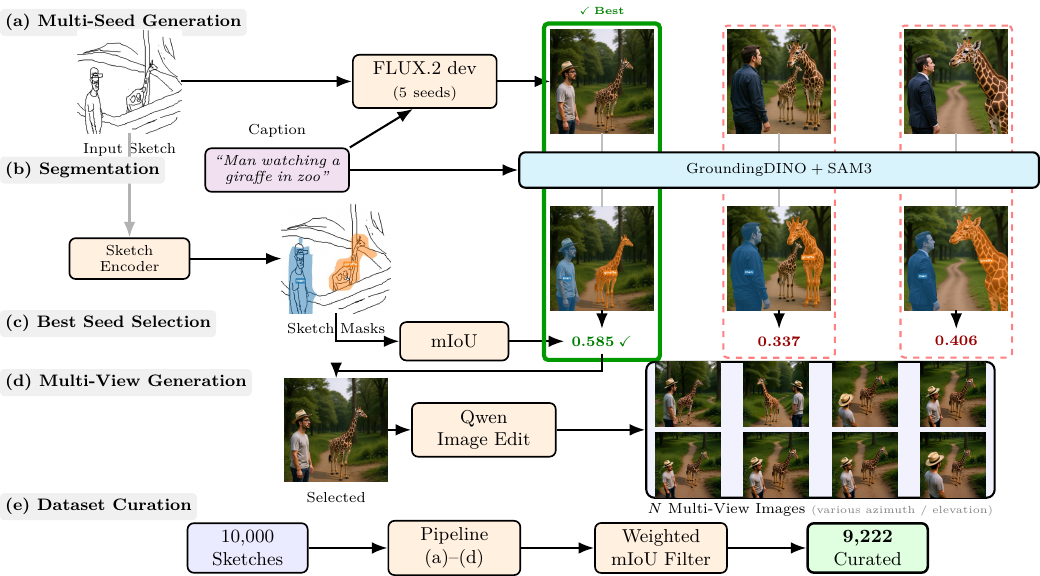}
  \caption{\textbf{S2WV dataset generation pipeline.}
  (a)~Multi-seed image generation from sketch and caption.
  (b)~Segmentation of sketch and generated images.
  (c)~Best seed selection via mIoU.
  (d)~Multi-view synthesis from the selected image.
  (e)~Dataset curation yields 9,222 samples. More details in \cref{sec:method:data}
    \vspace{-6mm}
  }
  \label{fig:dataset_pipeline}
\end{figure*}

No existing dataset pairs freehand scene sketches with geometrically consistent multi-view images. Unlike photographs, freehand sketches contain non-linear distortions that do not correspond to any standard 3D asset. We therefore propose a pipeline to bridge this domain gap through an automated generation-and-filtering process. The steps in our pipeline are summarized in~\cref{fig:dataset_pipeline}.

\begin{enumerate}[label=(\alph*)]
    \item \textbf{Multi-Seed Generation.}
For each of the ${\sim}$10{,}000 sketches in FS-COCO~\cite{chowdhury2022fs}, we synthesize five photorealistic candidate images using FLUX.2 dev~\cite{flux-2-2025}, conditioned on both the sketch and its original caption. 
    \item \textbf{Segmentation.} 
To ensure spatial fidelity, we perform semantic grounding by segmenting both the input sketch and the candidates. Sketch masks are extracted via a specialized sketch encoder~\cite{bourouis2024open}. For the generated images, we use spaCy to extract object nouns from the caption, which serve as prompts for zero-shot segmentation via GroundingDINO~\cite{liu2024grounding} and SAM3~\cite{carion2025sam}.

    \item \textbf{Best Seed Selection.} 
For each sketch, we select the candidate image $I^*$ that best aligns with the sketch's intended spatial layout. This is determined by maximizing the mean Intersection-over-Union (mIoU) between the sketch segmentation masks $\mathcal{M}_s$ and the image segmentation masks $\mathcal{M}_g$:
\begin{equation}
    I^* = \arg\max_{i \in \{1..5\}} \text{mIoU}(\mathcal{M}_s, \mathcal{M}_{g,i}).
\end{equation}

    \item \textbf{Multi-View Generation.} 
The selected frontal view is processed by Qwen Image Edit Angles~\cite{dx8152_qwen_edit_2509_multiple_angles_2025} to synthesize $N$ multi-view images at varying azimuth $\theta$ and elevation $\phi$. 

    \item \textbf{Dataset Curation.} 
We apply a minumum threshold on the mIoU to remove samples where the subjects are poorly aligned. 

\end{enumerate}

\noindent The result of our pipeline is a final corpus of \textbf{9\,222 high-quality samples}.

\subsection{Camera-Aware Attention Adapters (CA3)}
\label{sec:method:camera}

A video diffusion model has no intrinsic notion of camera viewpoint as its self-attention treats all tokens identically regardless of their 3D spatial relationship. For photograph-conditioned generation, dense appearance cues in the input can partially compensate; for freehand sketches, which provide only sparse, geometrically distorted strokes, the model receives no implicit 3D signal whatsoever. We therefore equip the model with an explicit geometric inductive bias through lightweight Camera-Aware Attention Adapters (CA3). Below we describe the construction of virtual camera matrices, the projective position encoding that consumes them, the adapter architecture, and a complementary domain-adaptation strategy.

\noindent \textbf{Virtual Camera Matrix Construction.}
\label{sec:camera_prope}
Since our multi-view images are generated via image editing rather than 3D rendering, ground-truth camera parameters are unavailable.
We construct virtual cameras by placing each viewpoint on a sphere of fixed radius $d$ around the scene origin. For a view at azimuth $\theta$ and elevation $\phi$, we compose a rotation $\mathbf{R}^{(n)} = \mathbf{R}_y(\theta) \cdot \mathbf{R}_x(\phi)$ and set the camera position to $\mathbf{t}^{(n)} = \mathbf{R}^{(n)} [0, 0, d]^\top$. 
The camera-to-world extrinsic is then:
\begin{equation}
    \mathbf{T}^{(n)} = \begin{bmatrix} \mathbf{R}^{(n)} & \mathbf{t}^{(n)} \\ \mathbf{0}^\top & 1 \end{bmatrix} \in SE(3).
\end{equation}
The intrinsic matrix $\mathbf{K}$ is shared across all views, constructed from a $60^{\circ}$ horizontal field-of-view with the principal point at the image centre.
These virtual matrices provide CA3 with a consistent geometric coordinate system, enabling viewpoint reasoning in the absence of ground-truth calibration

\noindent \textbf{Viewpoint Awareness via PRoPE.}
We integrate Projective Rotary Position Encoding (PRoPE)~\cite{li2025cameras}, which modulates attention using the scene's epipolar geometry rather than pixel coordinates alone. For each target view $n$, we construct a projective matrix from $\mathbf{T}^{(n)}$ and $\mathbf{K}$:
\begin{equation}
    \mathbf{P}^{(n)} = \begin{bmatrix} \mathbf{K} & \mathbf{0} \\ \mathbf{0}^\top & 1 \end{bmatrix} \cdot (\mathbf{T}^{(n)})^{-1}.
\end{equation}
By partitioning the attention head dimensions into projective and spatial subspaces, PRoPE ensures that the affinity between any two tokens is conditioned on their relative projective transform $\mathbf{P}^{(n_q)}\mathbf{P}^{(n_k)^{-1}}$, embedding epipolar constraints directly into the attention computation.

\noindent \textbf{CA3 Architecture.}
To preserve the powerful generative priors of the pretrained backbone, we implement CA3 as a parallel branch alongside the attention layers as illustrated in \cref{fig:method_overview}-(b). For each transformer block $\ell$, the hidden state $\mathbf{h}_\ell$ is updated as:
\begin{equation}
    \mathbf{h}'_\ell = \mathbf{h}_\ell + \mathcal{A}^{\text{self}}_\ell(\mathbf{h}_\ell) + \mathcal{A}^{\text{cam}}_\ell(\mathbf{h}_\ell, \{\mathbf{P}^{(n)}\}_{n=1}^N),
\end{equation}
where $\mathcal{A}^{\text{self}}_\ell(.)$ is LoRA-adapted backbone self-attention and $\mathcal{A}^{\text{cam}}_\ell(.,.)$ is our trainable camera adapter that uses the projective matrices of the views. To maintain computational efficiency, each adapter employs an $8\times$ dimensionality bottleneck ($1536 \to 192$) and utilizes only $H'{=}2$ attention heads, and its output projection is zero-initialised~\cite{zhang2023adding} to preserve the pretrained prior at the start of training.
The adapter architecture above addresses \emph{where} to attend across views; however, the backbone features themselves must also be adapted to the sketch domain.

\noindent \textbf{Sketch-Domain Adaptation via LoRA.}
\label{sec:lora_adaptation}
The video backbone is pretrained exclusively on natural videos; freehand sketches represent a significant domain shift that the camera adapters alone cannot bridge. We therefore apply LoRA~\cite{hu2022lora} with rank 16 to the query, key, value, and output projections of the backbone's self-attention layers.
While CA3 provides explicit geometric viewpoint control via PRoPE, LoRA handles the complementary task of adapting the backbone's feature representations for sketch-conditioned generation. Training follows a curriculum that lets domain adaptation stabilise before geometric supervision is introduced (\cref{sec:method:corr}). We validate this design choice in \cref{sec:ablations}.

\subsection{Correspondence Supervision Loss (CSL)}
\label{sec:method:corr}

\begin{figure*}[t]
  \centering
  \includegraphics[width=0.75\textwidth]{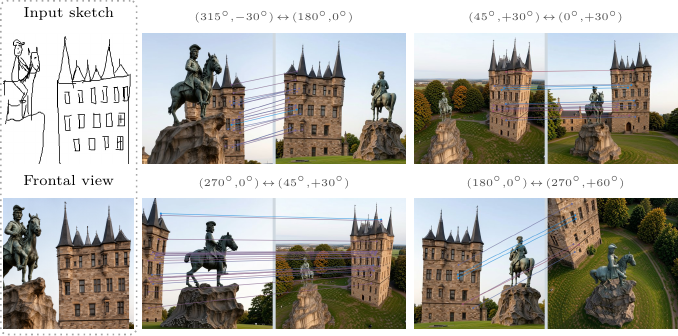}
  \caption{\textbf{SfM correspondences across views.}
  Each correspondence image shows two views with colored lines connecting matched keypoints identified by Structure-from-Motion (COLMAP).
  The view angles (azimuth, elevation) for each pair are indicated above the correspondence visualizations.
  These correspondences serve as supervision targets for our camera-aware attention adapters (\cref{sec:method:corr}).}
  \vspace{-6mm}
  \label{fig:colmap_correspondences}
\end{figure*}

The camera adapters give the model the \emph{capacity} to distinguish viewpoints, but this geometric inductive bias alone does not teach the model \emph{which} spatial locations across views correspond to the same 3D point. Without explicit supervision, the adapter's attention remains diffuse and spatially unstructured, as we verify empirically in \cref{sec:exp_geometric}. We therefore introduce CSL derived from SfM that directly teaches cross-view geometric reasoning (\cref{fig:colmap_correspondences}).

\noindent \textbf{Correspondence Extraction.}
Given the pseudo ground-truth multi-view images for each training sample, we run COLMAP~\cite{schonberger2016structure} to obtain sparse 3D reconstructions. For each reconstructed 3D point $\mathbf{X}_j$, COLMAP provides its projection coordinates $(u^{(n)}_j, v^{(n)}_j)$ in each observing view $n$, along with a confidence score $c^{(n)}_j$ derived from reprojection error. 
We construct pairwise supervision targets:
$$\mathcal{C} = \{(q_i, k_i, w_i)\}_{i=1}^M,$$

\noindent where $M$ is the total number of correspondence pairs, $q_i = (n_q, s_q)$ and $k_i = (n_k, s_k)$ specify the (view index, spatial patch index) for each pair and $w_i = \min(c^{(n_q)}_j, c^{(n_k)}_j)$ is the confidence weight.



\noindent \textbf{Sparse InfoNCE loss.}
Directly supervising the full attention matrix $\mathbf{A}^{(\ell)} \in \mathbb{R}^{T \times T}$ is memory-prohibitive ($T > 29{,}000$). We instead formulate correspondence supervision as a contrastive objective on the adapter's query and key projections. For each correspondence pair $(q_i, k_i)$, we extract query and key embeddings $\mathbf{q}_i$ and $\mathbf{k}_i$ from the CA3 module at a subset of layers $\mathcal{U} = \{10, 15, 20, 25\}$, corresponding to middle-to-deep layers where geometric reasoning is most pronounced. The sparse InfoNCE loss is:

$$\mathcal{L}_{\text{corr}} = -\frac{1}{M} \sum_{i=1}^{M} w_i \cdot \log \frac{\exp(\mathbf{q}_i^\top \mathbf{k}_i^+ / \tau)}{\exp(\mathbf{q}_i^\top \mathbf{k}_i^+ / \tau) + \sum_{j=1}^{N_{\text{neg}}} \exp(\mathbf{q}_i^\top \mathbf{k}_j^- / \tau)},$$

\noindent where $\mathbf{k}_i^+$ is the positive (corresponding) key, $\{\mathbf{k}_j^-\}_{j=1}^{N_{\text{neg}}}$ are negative keys sampled from non-corresponding token positions, and $\tau$ is a temperature parameter. This formulation has complexity $O(M \times N_{\text{neg}})$ rather than $O(T^2)$, reducing memory from 14\,GB to ${\sim}$1\,MB.

\noindent \textbf{Total training objective.}
The final training objective combines the flow matching loss with correspondence supervision:

$$\mathcal{L}_{\text{total}} = \mathcal{L}_{\text{flow}} + \lambda_{\text{corr}} \cdot \mathcal{L}_{\text{corr}},$$

\noindent where $\lambda_{\text{corr}}$ follows a curriculum schedule: it is held at zero during an initial warmup phase, then linearly ramped to its target value of $0.01$. This allows the flow matching loss and LoRA domain adaptation to stabilise before correspondence supervision is introduced.

\section{Experiments}
\label{sec:experiments}

We evaluate our method on sketch-to-multiview generation, assessing both per-view image quality and cross-view geometric consistency. We describe the experimental setup (\cref{sec:exp_setup}), compare against two strong baselines (\cref{sec:exp_comparison}), present ablations (\cref{sec:ablations}), visualise learned attention (\cref{sec:exp_geometric}), and test generalisation to unseen sketch domains (\cref{sec:exp_generalisation}).

\subsection{Experimental Setup}
\label{sec:exp_setup}

\noindent \textbf{Dataset and splits.}
We train on our S2MV dataset described in \cref{sec:method:data} with a total of 9{,}222 sketch--multiview samples ($N=33$ views each). We split these into 8{,}304 training, 441 validation, and 477 test samples, following \cite{bourouis2024data}. All reported metrics are computed on the test set across 3 seeds.

\noindent \textbf{Evaluation metrics.}
We report standard per-view image quality metrics against ground-truth views: peak signal-to-noise ratio (\textbf{PSNR}$\uparrow$), structural similarity index (\textbf{SSIM}$\uparrow$)~\cite{wang2004image}, and learned perceptual image patch similarity (\textbf{LPIPS}$\downarrow$)~\cite{zhang2018unreasonable} using an AlexNet backbone. To measure distributional realism, we report \textbf{FID}$\downarrow$~\cite{heusel2017gans} computed over all generated views versus all ground-truth views.
For semantic fidelity, we compute \textbf{CLIP-I}$\uparrow$ (cosine similarity between CLIP ViT-B/32 embeddings of generated and GT views). We additionally report \textbf{Corr-Acc}$\uparrow$, a geometric consistency metric that measures the percentage of COLMAP correspondences whose reprojection error on generated views falls below a threshold of 5 pixels; directly evaluating whether the model preserves the 3D structure visible in ground truth. All per-view metrics are averaged across views, seeds and test samples.

\noindent \textbf{Baselines.}
No prior method generates geometrically consistent multi-view output from a single freehand scene sketch. Therefore, we evaluate two state-of-the-art novel-view synthesis methods adapted to our setting via a shared two-stage pipeline: FLUX.2-dev~\cite{flux-2-2025} first translates each sketch to a photorealistic front view, then either \textbf{SEVA}~\cite{zhou2025stable} (Pl\"ucker-conditioned latent diffusion, $576 \times 576$) or \textbf{ViewCrafter}~\cite{yu2024viewcrafter} (point-cloud-guided video diffusion via DUSt3R~\cite{wang2024dust3r}, $576 \times 1024$) generates the remaining views. Full baseline configurations are provided in the supplementary material (\cref{sec:supp_baselines}).

\noindent \textbf{Implementation details.}
We set $N{=}33$ views at $480 \times 480$ resolution, building on the Wan\,2.1 video DiT (1.3B parameters)~\cite{wan2025wan} with frame replication ($N \to 4N{+}1$). We train on $8 \times$ NVIDIA A100 GPUs with AdamW~\cite{loshchilov2017decoupled}, LoRA~\cite{hu2022lora} rank 16 (${\sim}$5.9M parameters) alongside ${\sim}$35.4M CA3 parameters. The correspondence loss follows a curriculum schedule, ramping $\lambda_{\text{corr}}$ from $0$ to $0.01$. Extended training details are provided in the supplementary material (\cref{sec:supp_impl}).

\subsection{Comparison with Baselines}
\label{sec:exp_comparison}

\noindent \textbf{Quantitative results.}
\cref{tab:main_comparison} presents the main comparison across all metrics. Our single-stage method outperforms both two-stage baselines on five of six metrics despite operating directly from sketches without an intermediate photorealistic reference. The improvement is most pronounced on realism (FID $18.49$ vs.\ $46$--$48$) and geometric consistency (Corr-Acc $0.199$ vs.\ $0.161$/$0.136$), confirming that CA3 and CSL produce views that are individually plausible and mutually coherent. Perceptual quality likewise favours our method (LPIPS $0.632$ vs.\ $0.705$/$0.737$), as does semantic alignment with ground truth (CLIP-I $0.828$ vs.\ $0.756$/$0.773$). ViewCrafter achieves the highest SSIM ($0.338$), which we attribute to its tendency to produce smoother, lower-frequency outputs that score well on pixel-level structural similarity despite worse perceptual fidelity.

Our single-stage design also provides a $42\times$ inference speedup over ViewCrafter and $3.7\times$ over SEVA. More details in the supplementary material (\cref{sec:supp_efficiency}).

\begin{table*}[t]
\centering
\caption{\textbf{Quantitative comparison on S2MV test set} (477 samples, $N{=}33$ views each, 3 seeds). 
Our single-stage method is compared against two-stage baselines that first translate the sketch to a photograph via FLUX.2-dev.}
\label{tab:main_comparison}
\vspace{-2mm}
\resizebox{\textwidth}{!}{%
\begin{tabular}{l c c c c c c c c}
\toprule
\textbf{Method} & \textbf{\#Stages} & \textbf{Time/sample} & \textbf{PSNR}$\uparrow$ & \textbf{SSIM}$\uparrow$ & \textbf{LPIPS}$\downarrow$ & \textbf{FID}$\downarrow$ & \textbf{CLIP-I}$\uparrow$ & \textbf{Corr-Acc}$\uparrow$ \\
\midrule
SEVA~\cite{zhou2025stable} & 2-stage & ${\sim}$3.1\,min & 11.310 & 0.265 & 0.705 & 46.34 & 0.756 & 0.161 \\
ViewCrafter~\cite{yu2024viewcrafter} & 2-stage & ${\sim}$35\,min & 11.148 & 0.338 & 0.737 & 48.22 & 0.773 & 0.136 \\
\midrule
Ours & 1-stage & ${\sim}$50\,s & \textbf{12.169} & 0.302 & \textbf{0.632} & \textbf{18.49} & \textbf{0.828} & \textbf{0.199} \\
\bottomrule
\end{tabular}%
}
\end{table*}

\begin{figure*}[tp]
  \centering
  \includegraphics{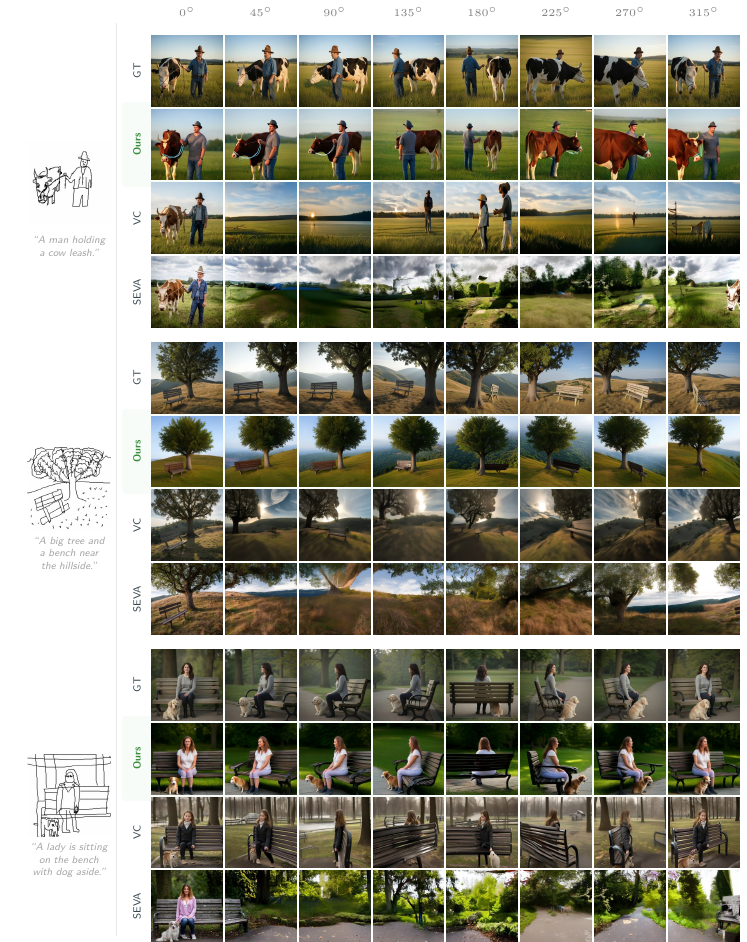}
    \vspace{-4mm}  
  \caption{\textbf{Qualitative comparison of multi-view generation from sketches.} For each example we show the input sketch with its text prompt (left) and eight azimuth views ($0^{\circ}$--$315^{\circ}$ at $45^{\circ}$ intervals) generated by each method. \textbf{GT}: ground-truth views; \textbf{Ours}: our single-pass sketch-to-multiview method; \textbf{VC}: ViewCrafter; \textbf{SEVA}: Stable Video Diffusion. Additional comparisons are provided in the supplementary material (\cref{sec:supp_qualitative}).
    \vspace{-6mm}
  }
  \label{fig:qualitative_comparison}
\end{figure*}

\noindent \textbf{Qualitative comparison.}
\cref{fig:qualitative_comparison} presents visual comparisons for representative test sketches. We visualize more examples in the supplementary material (\cref{sec:supp_qualitative}).

\subsection{Comparison to a Two-Stage FLUX$+$Qwen Pipeline}
\label{sec:exp_flux_qwen}

Our data pipeline (\cref{sec:method:data}) decomposes sketch-to-multiview into two off-the-shelf stages: FLUX.2-dev~\cite{flux-2-2025} (sketch$\to$frontal) followed by Qwen Image Edit Angles~\cite{dx8152_qwen_edit_2509_multiple_angles_2025} (frontal$\to$$N$ views). To test whether our trained single-stage model is merely a faster substitute for this pipeline, we evaluate a strong two-stage \textbf{FLUX$+$Qwen} variant on the S2MV test set (\cref{tab:flux_qwen}) averaged across 3 different seeds.

\begin{table}[t]
\centering
\caption{\textbf{Two-stage FLUX$+$Qwen vs.\ ours} on S2MV. ``Frontal'' evaluates the single front view (mIoU $=$ sketch fidelity); ``Multi-view'' evaluates all $N$ views. Ours wins on both stages despite seeing only a sketch at inference.}
\label{tab:flux_qwen}
\vspace{-2mm}
\resizebox{\textwidth}{!}{%
\begin{tabular}{l cccc ccc}
\toprule
& \multicolumn{4}{c}{\textbf{Frontal}} & \multicolumn{3}{c}{\textbf{Multi-view} ($N{=}33$)} \\
\cmidrule(lr){2-5}\cmidrule(lr){6-8}
\textbf{Method} & \textbf{FID}$\downarrow$ & \textbf{CLIP-I}$\uparrow$ & \textbf{LPIPS}$\downarrow$ & \textbf{mIoU}$\uparrow$ &
   \textbf{FID}$\downarrow$ & \textbf{CLIP-I}$\uparrow$ & \textbf{Corr-Acc}$\uparrow$ \\
\midrule
FLUX$+$Qwen (3 seeds) & 47.64 & 0.776 & 0.731 & 0.338 & 51.77 & 0.804 & 0.174 \\
Ours & \textbf{36.33} & \textbf{0.781} & \textbf{0.722} & \textbf{0.353} & \textbf{19.12} & \textbf{0.822} & \textbf{0.191} \\
\bottomrule
\end{tabular}%
}
\vspace{-2mm}
\end{table}

Despite seeing strictly less information at inference, a sketch rather than a curated photograph, our model wins on \emph{every} metric, even surpassing the pipeline on \emph{frontal} sketch fidelity (mIoU $0.353$ vs.\ $0.338$) and realism (FID $36.33$ vs.\ $47.64$); the gap widens for multi-view. The bottleneck is thus the sketch$\to$frontal step, which our five-seed mIoU curation (\cref{sec:method:data}) closes during training. The LoRA branch is therefore \emph{not} a speed-only convenience: it absorbs the curated S2MV statistics, so inference needs no FLUX seed search. The same argument applies to any sketch$\to$image$\to$multi-view decomposition (\eg a sketch-ControlNet front-end), of which FLUX$+$Qwen is a strong public instance.

\subsection{Ablation Study}
\label{sec:ablations}
We ablate each component under identical training settings and report results on the S2MV test set averaged across 3 seeds (\cref{tab:ablation}).

\begin{table*}[t]
\centering
\caption{\textbf{Ablation study.} We remove each contribution independently. Row (a) is the full model. Removing any single component degrades both per-view quality and geometric consistency, confirming that the contributions are mutually reinforcing.}
\label{tab:ablation}
\vspace{-2mm}
\resizebox{\textwidth}{!}{%
\begin{tabular}{c l c c c c c c c}
\toprule
& \textbf{Configuration} & \textbf{PSNR}$\uparrow$ & \textbf{SSIM}$\uparrow$ & \textbf{LPIPS}$\downarrow$ & \textbf{FID}$\downarrow$ & \textbf{CLIP-I}$\uparrow$ & \textbf{Corr-Acc}$\uparrow$ & \textbf{T. params} \\
\midrule
(a) & Full model (CA3 $+$ CSL $+$ LoRA) & \textbf{12.169} & 0.302 & \textbf{0.632} & \textbf{18.49} & \textbf{0.828} & \textbf{0.199} & ${\sim}$41.3M \\
\midrule
(b) & w/o CSL ($\lambda_{\text{corr}}{=}0$) & 12.073 & 0.287 & 0.664 & 20.37 & 0.817 & 0.175 & ${\sim}$41.3M \\
(c) & w/o CA3 (LoRA only) & 5.026 & 0.266 & 0.819 & 266.06 & 0.632 & 0.183 & ${\sim}$5.9M \\
(d) & w/o LoRA (CA3 only) & 12.211 & \textbf{0.304} & 0.644 & 19.27 & 0.823 & 0.188 & ${\sim}$35.4M \\
(e) & w/o frame replication & 12.198 & \textbf{0.304} & 0.652 & 42.54 & 0.786 & 0.174 & ${\sim}$41.3M \\
\bottomrule
\end{tabular}%
}
\end{table*}

\begin{itemize}

\item \textbf{CSL} (\cref{tab:ablation}b).
Removing the correspondence loss ($\lambda_{\text{corr}}{=}0$) retains geometric capacity but removes explicit supervision. Row~(b) shows the largest Corr-Acc drop among non-catastrophic ablations ($0.175$ vs.\ $0.199$) and degraded LPIPS ($0.664$ vs.\ $0.632$), confirming that \textbf{CSL} is essential for translating adapter capacity into cross-view consistency.

\item \textbf{CA3} (\cref{tab:ablation}c).
Without our CA3 camera adapter, the model has no mechanism to distinguish viewpoints. Row~(c) is nearly catastrophic: PSNR drops to $5.026$, FID degrades to $266.06$, and CLIP-I drops to $0.632$, confirming that LoRA alone is insufficient for the multi-view setting.

\item \textbf{LoRA} (\cref{tab:ablation}d).
Row~(d) is competitive on per-view quality (PSNR $12.211$, FID $19.27$), suggesting our CA3 carry most of the representational burden. However, Corr-Acc drops to $0.188$, indicating that LoRA's domain adaptation complements the adapters for fine-grained geometric consistency.

\item \textbf{Frame replication}  (\cref{tab:ablation}e).
Without frame replication, the VAE's $4\times$ temporal compression merges neighbouring views into shared latent frames. Per-view PSNR remains competitive ($12.198$), but FID more than doubles ($42.54$ vs.\ $18.49$) and Corr-Acc degrades to $0.174$, confirming that preserving view identity through the VAE bottleneck is critical.

\end{itemize}

Taken together, all components are mutually reinforcing: removing any one degrades geometric consistency.


\subsection{Geometric Consistency Evaluation}
\label{sec:exp_geometric}

\begin{figure*}[t]
  \centering
  \includegraphics{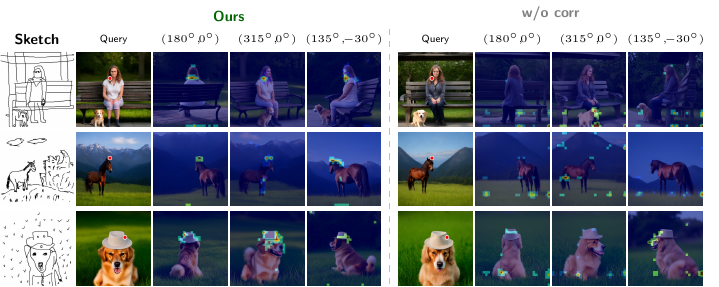}
  \caption{\textbf{Attention correspondence visualisation.} Query pixel (red dot) in the front view; heatmaps show attention at layer~20 over three target viewpoints. Left: with correspondence supervision. Right: without (ablation \cref{tab:ablation} row~b).
    \vspace{-6mm}
  }
  \label{fig:attention_correspondence}
\end{figure*}

To understand \emph{how} correspondence supervision shapes attention, we visualise camera attention maps at layer~20 (\cref{fig:attention_correspondence}). Given a query pixel in the front view, we extract attention weights over three target views at varying viewpoints. With correspondence supervision, attention concentrates sharply on the geometrically correct region, even for the challenging $180^{\circ}$ back view. Without it (ablation row~b), attention is diffuse and spatially unstructured, confirming that CSL directly teaches the adapters where to attend.

\subsection{Generalisation to Unseen Sketch Domains}
\label{sec:exp_generalisation}

\begin{figure*}[t]
  \centering
  \includegraphics{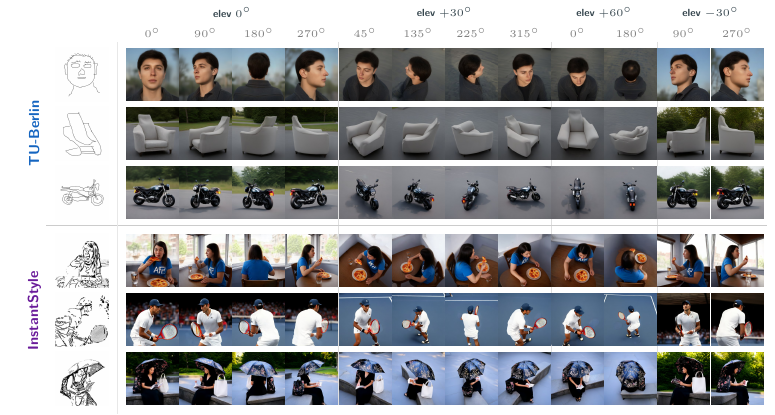}
      \vspace{-6mm}
  \caption{\textbf{Generalisation to unseen sketch domains.} 12 views per sketch at four elevations. (Top)~TU-Berlin object sketches~\cite{eitz2012humans}: consistent multi-view output despite a distinctly different sketch style. (Bottom)~InkScenes~\cite{tang2025instance}: robust to dense, textured scene compositions unseen during training.
      \vspace{-6mm}
}
  \label{fig:generalisation}
\end{figure*}

Our model is trained exclusively on FS-COCO freehand sketches. To test whether the learned geometric reasoning transfers to unseen sketch distributions or not, we evaluate on sketches from two out-of-distribution datasets (see \cref{sec:supp_unseen_sketch} in supplementary) and the qualitative results are illustrated in~\cref{fig:generalisation}. From these results, it is clear that per-view image quality and consistency between multi-views are plausible, even if the data distributions are different than FS-COCO.

\subsection{Limitations}
\label{sec:limitations}
Our method is limited by (i)~training scale ($8\,304$ samples vs.\ $>1$M for baselines), (ii)~generated rather than rendered ground truth, and (iii)~$480 \times 480$ resolution; we discuss these in the supplementary material (\cref{sec:supp_limitations}).
\vspace{-2mm}

\section{Conclusion}

We presented the first method to generate geometrically consistent multi-view scenes from a single freehand sketch in one forward pass, unifying a curated S2MV dataset, lightweight parallel CA3, and a sparse CSL into three mutually reinforcing components.
Beyond multi-view synthesis, we believe this opens a broader direction: sketching as a lightweight, accessible interface for 3D content creation.
More generally, our work demonstrates that combining camera-aware attention adapters with explicit correspondence supervision is an effective paradigm for lifting 2D generative models into geometrically grounded multi-view generators.



%
%
\bibliographystyle{splncs04}
\bibliography{main}

\clearpage

\appendix
\renewcommand{\theHsection}{supp.\arabic{section}}
\renewcommand{\theHsubsection}{supp.\arabic{section}.\arabic{subsection}}
\renewcommand{\theHfigure}{supp.\arabic{figure}}
\renewcommand{\theHtable}{supp.\arabic{table}}
\section*{Supplementary Material}

This supplementary document provides additional details that complement the main paper. \cref{sec:supp_qualitative} presents additional qualitative comparisons. \cref{sec:supp_impl} gives extended implementation details including gradient flow, training configuration, and correspondence sampling. \cref{sec:supp_baselines} describes the baseline configurations used for evaluation. \cref{sec:supp_efficiency} reports inference efficiency comparisons. \cref{sec:supp_limitations} provides an extended discussion of limitations.

\medskip
\noindent \textbf{Interactive Gallery.}
We include an interactive gallery on the project page. Open it in any browser to browse 50 test samples side-by-side across methods (Ours, SEVA, ViewCrafter) compared to ground truth views. The gallery supports filtering by elevation ($0^\circ$, $\pm30^\circ$, $+60^\circ$) and displays the input sketch with its caption alongside the generated views.

\section{Additional Qualitative Comparisons}
\label{sec:supp_qualitative}

\begin{figure*}[t]
  \centering
  \includegraphics{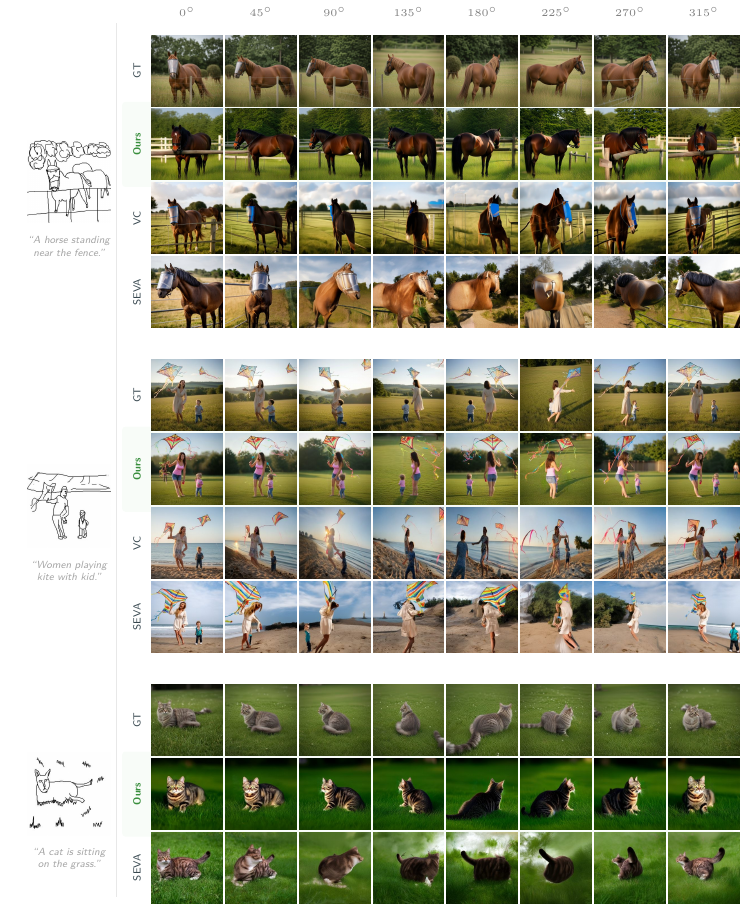}
  \caption{\textbf{Additional qualitative comparison.} Same format as Fig. 5: input sketch with text prompt (left) and eight azimuth views generated by each method.}
  \label{fig:supp_qualitative}
\end{figure*}

\begin{figure*}[t]
  \centering
  \includegraphics{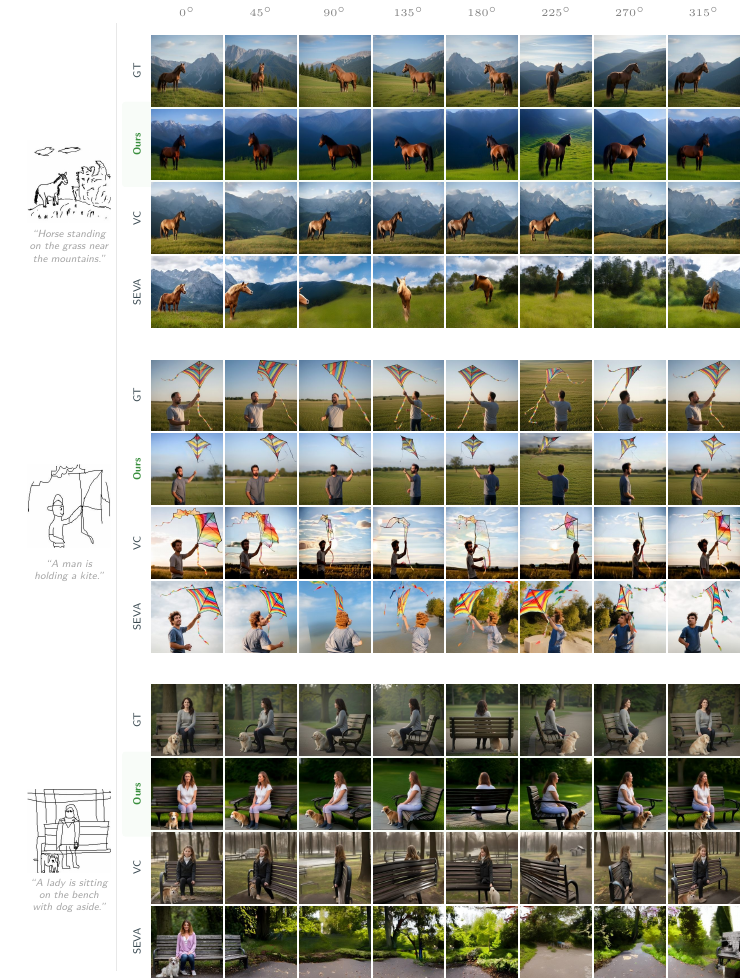}
  \caption{\textbf{Additional qualitative comparison.} Same format as Fig. 5: input sketch with text prompt (left) and eight azimuth views generated by each method.}
  \label{fig:supp_qualitative_2}
\end{figure*}

\cref{fig:supp_qualitative} and \cref{fig:supp_qualitative_2} presents additional visual comparisons between our method and both baselines on test set sketches, complementing Fig. 5 in the main paper.

\section{Extended Implementation Details}
\label{sec:supp_impl}

\noindent \textbf{Hidden state caching for gradient flow.}
A subtle but critical implementation detail concerns gradient propagation through the correspondence loss. The camera-aware attention adapter is integrated as a parallel branch in each DiT block:

$$\mathbf{h}'_\ell = \mathbf{h}_\ell + \mathcal{A}^{\text{self}}_\ell(\mathbf{h}_\ell) + \mathcal{A}^{\text{cam}}_\ell(\mathbf{h}_\ell),$$

\noindent where $\mathbf{h}_\ell$ is the input to the block. To compute the correspondence loss, we require access to the hidden states $\mathbf{h}_\ell$ that serve as input to the camera attention module $\mathcal{A}^{\text{cam}}_\ell$, specifically at the supervised layers $\ell \in \mathcal{L} = \{10, 15, 20, 25\}$.

We implement a hidden state caching mechanism that stores $\mathbf{h}_\ell$ during the forward pass \emph{without detaching} it from the computation graph. After the forward pass completes and the flow matching loss is computed, we retrieve these cached hidden states and compute the correspondence loss by projecting them through the same $\mathbf{W}_Q, \mathbf{W}_K$ matrices used in $\mathcal{A}^{\text{cam}}_\ell$:

\begin{align}
\mathbf{Q}^{(\ell)} &= \mathbf{W}_Q^{(\ell)} \mathbf{h}_\ell, \\
\mathbf{K}^{(\ell)} &= \mathbf{W}_K^{(\ell)} \mathbf{h}_\ell.
\end{align}

Since the cached $\mathbf{h}_\ell$ retains its gradient connection and we apply the projection matrices directly, the backward pass propagates gradients from the correspondence loss to both $\mathbf{W}_Q^{(\ell)}$ and $\mathbf{W}_K^{(\ell)}$, enabling effective learning of geometrically-aware attention patterns.

\noindent \textbf{PRoPE application in loss computation.}
After computing the raw query and key projections, we apply the same PRoPE transformations used during the forward pass:

\begin{align}
\tilde{\mathbf{Q}}^{(\ell)} &= \mathbf{D}_{\mathbf{Q}}^{(n)} \mathbf{Q}^{(\ell)}, \\
\tilde{\mathbf{K}}^{(\ell)} &= \mathbf{D}_{\mathbf{K}}^{(n)} \mathbf{K}^{(\ell)},
\end{align}

\noindent where $\mathbf{D}_{\mathbf{Q}}^{(n)}$ and $\mathbf{D}_{\mathbf{K}}^{(n)}$ are the PRoPE transformation matrices for view $n$. This ensures the correspondence supervision operates in the same geometric embedding space as the actual attention computation.

\noindent \textbf{Training configuration.}
We train on $8 \times$ NVIDIA A100 GPUs using the Accelerate library for distributed training. Each GPU processes one sample with gradient accumulation of 1 step, yielding an effective batch size of 8. We use AdamW~\cite{loshchilov2017decoupled} with differential learning rates: $10^{-4}$ for camera adapter parameters, $10^{-5}$ for LoRA parameters, with 600 warmup steps and cosine annealing. Gradient clipping is set to 1.0. LoRA~\cite{hu2022lora} with rank 16 is applied to the DiT attention layers (\texttt{q,k,v,o} projections), resulting in ${\sim}$5.9M trainable LoRA parameters alongside the ${\sim}$35.4M camera adapter parameters.

\noindent \textbf{Correspondence sampling.}
We sample up to $M = 10{,}000$ correspondence pairs per training sample. When the total number of available correspondences exceeds this limit, we uniformly subsample while preserving the confidence distribution. We use $N_{\text{neg}} = 128$ negative samples per query for the InfoNCE loss.


\section{Baseline Configurations}
\label{sec:supp_baselines}

Both baselines use a shared first stage: FLUX.2-dev~\cite{flux-2-2025} (4-bit quantised, ${\sim}$18B parameters) translates each test sketch to a photorealistic front view with the prompt ``\texttt{Convert to a photorealistic image of \{caption\}}'', using 28 denoising steps with guidance scale 4.0.

\noindent \textbf{SEVA (Stable Virtual Camera).}
\label{sec:impl_seva}
We use SEVA v1.1~\cite{zhou2025stable}, a 1.3B-parameter latent diffusion model that generates novel views conditioned on input images and target camera poses at $576 \times 576$ resolution. SEVA employs a context window of $T{=}21$ frames and a two-pass sampling strategy (\texttt{img2trajvid}): the first pass generates sparse anchor views uniformly spaced along the target trajectory, the second pass fills in remaining views conditioned on both the input and the generated anchors. We use default hyperparameters: 50 denoising steps, classifier-free guidance scale 2.0 (both passes), \texttt{MultiviewCFG} guider for the first pass, \texttt{MultiviewTemporalCFG} for the second, with a minimum guidance scale of 1.2.

To ensure a fair comparison, we generate views at the same $N$-view azimuth/elevation grid used by our method (8 azimuths $\times$ 4 elevations $+$ 1 front view; Sec. 3.3). Since our method uses an OpenGL camera convention (Y-up) while SEVA uses OpenCV (Y-down), we apply a coordinate flip $\mathbf{T}^{\text{cv}}_{c} = \mathbf{T}^{\text{gl}}_{c} \cdot \text{diag}(1, -1, -1, 1)$ to convert camera-to-world matrices. We use pixel-space pinhole intrinsics with a $54^{\circ}$ field of view (SEVA's default) and principal point at the image centre.

We run the full SEVA evaluation (477 test sketches $\times$ 3 seeds $=$ 1{,}431 generations) distributed across 16 GPUs. Each sample requires a FLUX forward pass (${\sim}$22\,s) followed by SEVA two-pass generation (${\sim}$3\,min).

\noindent \textbf{ViewCrafter.}
\label{sec:impl_viewcrafter}
ViewCrafter~\cite{yu2024viewcrafter} is a point-cloud-guided video diffusion pipeline that synthesises novel views in three steps: (a)~DUSt3R~\cite{wang2024dust3r} estimates a dense point cloud from the input image, (b)~the point cloud is rendered from the target camera via PyTorch3D~\cite{ravi2020pytorch3d}, producing a sparse rendering, and (c)~a video diffusion model (based on DynamiCrafter~\cite{xing2024dynamicrafter}) inpaints the missing regions. For each target view, ViewCrafter generates a 25-frame video interpolating from input to target camera; we extract the last frame as the novel-view image.

ViewCrafter uses a spherical coordinate system parameterised by angular offsets $d_\phi$ (azimuth) and $d_\theta$ (elevation) from the input view. We map our view grid as $d_\phi = \text{azim}$ and $d_\theta = -\text{elev}$ (negated to match conventions). For azimuths $> 180^\circ$, we take the shortest-path wraparound (\eg $225^\circ \to -135^\circ$). ViewCrafter operates at a fixed $576 \times 1024$ landscape resolution; we resize the $1024 \times 1024$ FLUX output to $1024 \times 576$ before DUSt3R and resize each output back to $576 \times 576$ after generation.

We use default hyperparameters: 50 DDIM steps, guidance scale 7.5, guidance rescale 0.7. DUSt3R uses ViT-Large at resolution 512 with 300 iterations of point cloud optimisation. Each sample requires one FLUX forward pass (${\sim}$30\,s) followed by $N{-}1$ ViewCrafter generations (${\sim}$65\,s each), for a total of ${\sim}$35\,min per sample.

\section{Unseen Sketch Datasets}
\label{sec:supp_unseen_sketch}

\noindent \textbf{TU-Berlin sketch dataset.}
The TU-Berlin dataset~\cite{eitz2012humans} contains single-object freehand sketches (250 categories, ${\sim}$20{,}000 drawings) with a distinctly different style from FS-COCO scene sketches.

\noindent \textbf{InkScenes (InstantStyle variant).}
We additionally evaluate on scene-level sketches from the InstantStyle split of the InkScenes dataset~\cite{tang2025instance}. These sketches were produced by applying the InstantStyle~\cite{wang2024instantstyle} style-transfer method to photographs from Visual Genome~\cite{krishna2017visual}, yielding expressive, natural-looking scene sketches across 72 object categories. The resulting sketches exhibit denser strokes, richer textures, and more complex multi-object compositions than FS-COCO.

\section{Inference Efficiency}
\label{sec:supp_efficiency}

\cref{tab:efficiency} compares inference cost across all methods. Our single-stage method generates all $N$ views in a single forward pass ($\sim$50\,s on a single A100 GPU), while SEVA requires a FLUX pass followed by a two-pass diffusion schedule ($\sim$3.1\,min), and ViewCrafter requires $N{-}1$ separate video diffusion runs after DUSt3R point cloud estimation ($\sim$35\,min) --- a $42\times$ slowdown. The efficiency gap stems from our architectural choice: by conditioning a video DiT on all $N$ camera poses simultaneously, we amortise the denoising cost across views rather than generating each independently.

\begin{table}[t]
\centering
\caption{\textbf{Inference efficiency.} Wall-clock time per sample on a single A100 GPU ($N{=}33$ views).}
\label{tab:efficiency}
\vspace{-2mm}
\begin{tabular}{l c c c}
\toprule
\textbf{Method} & \textbf{Passes} & \textbf{Time/sample} & \textbf{Speedup} \\
\midrule
FLUX$+$SEVA & 3 (FLUX$+$2$\times$SEVA) & ${\sim}$3.1\,min & $3.7\times$ slower \\
FLUX$+$ViewCrafter & $N$ (FLUX$+$$N{-}1$$\times$VC) & ${\sim}$35\,min & $42\times$ slower \\
Ours & 1 & ${\sim}$50\,s & --- \\
\bottomrule
\end{tabular}
\end{table}

\section{Extended Limitations Discussion}
\label{sec:supp_limitations}

\noindent \textbf{Dataset scale.}
Our training set comprises 8{,}304 training samples (9{,}222 total including validation and test), orders of magnitude smaller than the datasets used by baselines (SEVA is trained on $>1$M real-world video clips). This limits the diversity of scenes our model can faithfully reconstruct; complex multi-object arrangements with heavy occlusion sometimes produce blurred or hallucinated content in rear views ($180^{\circ}$ azimuth). Scaling the data pipeline (Sec. 3.2) to generate more training samples is a natural direction for improvement.

\noindent \textbf{Generated ground truth.}
Our multi-view training views are produced by an image editing model~\cite{dx8152_qwen_edit_2509_multiple_angles_2025}, not rendered from 3D assets. While the SfM filtering ensures geometric consistency, subtle artefacts (lighting shifts, texture drift) propagate into training. This is an inherent limitation of training multi-view generation without access to ground-truth 3D scenes. Our S2MV dataset presented in this work is a good starting point to enable sketching as an accessible interface for 3D content creation. 

\noindent \textbf{Resolution.}
We train at $480 \times 480$ resolution, constrained by GPU memory when processing $N$ views simultaneously ($4N{+}1$ frames after replication). Higher resolutions are feasible with gradient checkpointing and model parallelism; preliminary experiments retraining our full model at $576 \times 576$ already improve over the $480 \times 480$ configuration (PSNR $12.31$, FID $17.92$, Corr-Acc $0.205$), indicating that the quality ceiling is not architectural. We leave a full higher-resolution study to future work.

\end{document}